\documentclass{article}

 \usepackage[preprint]{neurips_2026}

\usepackage[utf8]{inputenc} 
\usepackage[T1]{fontenc}    
\usepackage{hyperref}       
\usepackage{url}            
\usepackage{booktabs}       
\usepackage{amsfonts}       
\usepackage{nicefrac}       
\usepackage{microtype}      
\usepackage{xcolor}         
\usepackage{amsmath}
\usepackage{graphicx}      
\usepackage[table]{xcolor} 
\usepackage{booktabs}      
\usepackage{multirow}      

\makeatletter
\NewEnviron{arxivhide}{
  \if@preprint
  \else
    \BODY
  \fi
}
\makeatother

\title{SeqRoute: Global Budget-Aware Sequential LLM Routing via Offline Reinforcement Learning}

\author{%
  Zhongling Xu \qquad Shunan Zheng \qquad Wei Wang \\
  Department of Operations Research and Industrial Engineering\\
  The University of Texas at Austin\\
  Austin, TX 78712 \\
  \texttt{xuzhl@austin.utexas.edu} \\
}

\begin{document}

\maketitle

\begin{abstract}
Existing LLM routing frameworks treat queries as independent events, neglecting the sequential nature of real-world user sessions constrained by global computational budgets. This mismatch inevitably leads to \textit{budget bankruptcy}: myopic routing policies exhaust resources on early interactions, forcing subsequent and often more complex queries onto inadequate models. We introduce \textbf{SeqRoute}, a framework that formulates multi-turn routing as a finite-horizon Markov Decision Process and solves it via offline reinforcement learning. By incorporating the remaining budget into the state space and training with Conservative Q-Learning (CQL), SeqRoute learns \textit{delayed gratification} to strategically preserve resources for high-stakes turns later in the session. To overcome data starvation, we propose \textit{Hindsight Budget Relabeling} (HBR). This technique retrospectively simulates historical trajectories under diverse hypothetical budgets, expanding 10,000 raw sessions into 2.38 million transitions enriched with critical bankruptcy signals. At deployment, a dynamic $\lambda$-sweep mechanism enables zero-shot navigation of the cost-quality Pareto frontier without retraining. Extensive evaluations demonstrate that SeqRoute reduces operational costs by 6.0--73.5\% while maintaining or improving quality, and suppresses bankruptcy rates to under 1\%, strictly dominating behavior cloning, budget-aware heuristics, and static baselines across the entire Pareto frontier.
\end{abstract}

\section{Introduction}
\label{sec:intro}

Large Language Models (LLMs) have achieved remarkable success across diverse
applications, but deploying state-of-the-art (SOTA) models incurs prohibitive
inference costs. LLM routing has emerged as a pragmatic paradigm that dynamically
assigns incoming queries to either cost-effective smaller models or highly capable
larger models, optimizing the trade-off between cost and
performance~\cite{chen2023frugalgpt, ong2025routellm, ding2024hybridllm}.
A rich body of routing strategies now exists, leveraging preference
data~\cite{ong2025routellm}, confidence scores~\cite{chuang2025confidence},
graph-based representations~\cite{feng2025graphrouter}, and adaptive test-time
compute~\cite{ding2025bestroute, zhang2025stick, pan2025rtr}. These advances are
supported by open benchmarks for systematic comparison~\cite{feng2025llmrouter,
hu2024routerbench, lu2026routerarena, huang2025routereval}.

Despite this progress, we identify a fundamental mismatch between how routing
is widely studied and how it is actually deployed: \textbf{existing frameworks treat
each query as an independent event, yet real-world users engage in multi-turn
sessions constrained by a global computational or financial budget}. These two
settings require qualitatively different decision-making. Consider a user session
with a fixed budget spanning four conversation turns. A myopic router, though optimal
for independent queries, may encounter an early simple question and confidently route
it to the expensive 70B model. By the third turn, when the user asks a genuinely
hard follow-up question, the budget is already depleted: the router either fails entirely or
falls back to a weak model that produces an inadequate answer. We term this systemic
failure mode \emph{budget bankruptcy}. A router that cannot reason about future
consequences of today's spending will inevitably exhaust resources prematurely,
collapsing exactly when capability matters most. Empirically, we show that greedy
behavior cloning suffers a bankruptcy rate exceeding 31.8\% under realistic session
budgets, which is consistently worse than our SeqRoute agent across all budget levels,
demonstrating that no amount of single-turn optimization can
prevent budget bankruptcy.

This reveals that session-level routing is not merely a harder variant of per-query
routing but a \textbf{qualitatively different problem}. The core challenge shifts
from per-query difficulty estimation to \emph{sequential planning under resource
constraints}: each model-selection decision changes what remains affordable for
future turns. We therefore propose \textbf{SeqRoute}, which formalizes multi-turn
LLM routing as a finite-horizon Markov Decision Process (MDP) whose state explicitly
encodes both the conversational context and the \emph{remaining session budget}.
By treating model selection as a long-horizon planning problem, SeqRoute's agent
learns \emph{delayed gratification} to deliberately route early queries to
cost-effective models to preserve resources for high-stakes turns later in the session.

Moving from formulation to implementation presents two obstacles. \textbf{First, data
starvation}: standard interaction logs lack budget annotations, depriving the agent
of the bankruptcy signals essential for learning constraint satisfaction. To
circumvent this, we introduce \textit{Hindsight Budget Relabeling} (HBR). Inspired
by Hindsight Experience Replay~\cite{andrychowicz2017hindsight}, HBR retrospectively
simulates recorded trajectories under a spectrum of hypothetical initial budgets.
Without incurring any additional API costs, HBR projects unconstrained rollouts into
a massive dataset of over 2.38 million transitions rich with bankruptcy signals. We
prove that this synthesis is theoretically valid: because conversation dynamics are
conditionally independent of the budget state, splicing a deterministic budget
transition into unconstrained logs yields unbiased trajectories of the augmented MDP.
\textbf{Second, safety}: online exploration is financially prohibitive and risks
catastrophic user experience. We instantiate SeqRoute with Conservative Q-Learning
(CQL)~\cite{kumar2020conservative}, which learns a lower-bound Q-function by
explicitly penalizing out-of-distribution (OOD) actions. This conservative penalty
systematically prevents the agent from invoking expensive models in budget-impoverished
states where historical data lacks evidence of survival, fundamentally distinguishing
SeqRoute from imitation learning approaches that naively mimic training distributions
and lack any fail-safe mechanism.

Finally, deployment budgets fluctuate across users, applications, and time. Rather than
training separate policies for each budget level, we introduce a \textbf{$\lambda$-sweep}
mechanism grounded in Lagrangian duality. By shifting the cost penalty to test-time
inference through $a_t^* = \arg\max_a [Q(s_t, a) - \lambda \cdot c(a)]$, a single SeqRoute
policy continuously navigates the full cost-quality Pareto frontier zero-shot, enabling
smooth adaptation to varying economic constraints without retraining.

The main contributions of this work are:
\begin{itemize}
    \item \textbf{The Session-Budget MDP.} We formally define budget-constrained
    sequential routing as a finite-horizon MDP with an augmented state space that
    explicitly tracks remaining budget, and introduce \textbf{SeqRoute} as a principled
    offline RL framework for solving it.

    \item \textbf{Hindsight Budget Relabeling (HBR).} We propose HBR to overcome
    offline RL data starvation. By sweeping hypothetical budget constraints across
    recorded trajectories, HBR expands standard session logs into over 2.38 million
    bankruptcy-annotated transitions at zero additional cost. We provide a theoretical
    validity guarantee for this synthesis.

    \item \textbf{Zero-Shot Pareto Navigation and Emergent Behavior.} We introduce the
    $\lambda$-sweep mechanism, enabling a single SeqRoute policy to continuously trace
    the cost-quality Pareto frontier at deployment. Extensive evaluations demonstrate
    that SeqRoute strictly dominates all baselines, and micro-trajectory analyses confirm
    the agent genuinely learns \emph{delayed gratification}, suppressing expensive model
    usage in budget-tight early turns, then strategically deploying it on decisive final
    queries.
\end{itemize}

\section{Related Work}
\label{sec:related_work}

The rapid proliferation of Large Language Models (LLMs) with varying capabilities and inference costs has driven the need for intelligent routing mechanisms~\cite{hu2024routerbench, huang2025routereval, lu2026routerarena, feng2025llmrouter}. Our work builds upon and extends two primary lines of research: LLM Routing Strategies and Offline Reinforcement Learning.

\subsection{Evolution of LLM Routing Strategies}

\textbf{Single-Turn Routing.} Early approaches to LLM routing focused on per-query model selection to balance response quality and compute. Cascade-style methods and preference-based classifiers, such as FrugalGPT~\cite{chen2023frugalgpt} and RouteLLM~\cite{ong2025routellm}, demonstrated that reserving stronger models for harder instances yields substantial cost savings. Subsequent work improved query-model representations through confidence-aware signals~\cite{chuang2025confidence}, latent preference modeling~\cite{zhang2025stick}, heterogeneous graph structures~\cite{feng2025graphrouter}, and knowledge-graph-guided multi-agent collaboration~\cite{zhang2025agentrouter}. For dynamic query streams, contextual-bandit frameworks like MixLLM~\cite{wang2025mixllm} and RadialRouter~\cite{jin2025radialrouter} introduced adaptive model selection. Extensions to retrieval-augmented generation~\cite{zhang2025ragrouter} and in-context learning~\cite{wang2025iclrouter} further enriched the routing context.

Despite this progress, these methods share a common limitation: they treat each query in isolation. A router optimized for per-query accuracy has no mechanism to reason about the cumulative cost of its decisions across a multi-turn session.

\textbf{Toward Budget-Aware and Sequential Routing.} As deployment costs scaled, the field began to incorporate explicit budget constraints. At the single-model level, SelfBudgeter~\cite{li2025selfbudgeter} and BudgetThinker~\cite{wen2025budgetthinker} adaptively control token budgets, while CP-Router~\cite{su2026cprouter} and Route to Reason~\cite{pan2025rtr} selectively invoke heavy reasoning strategies based on uncertainty. Concurrently, routing began transitioning toward multi-turn and multi-agent orchestration. GMTRouter~\cite{xie2025gmtrouter} models personalized routing from sparse multi-turn histories, while RouteMoA~\cite{wang2026routemoa} and AdaptEvolve~\cite{ray2026adaptevolve} dynamically optimize model selection within mixture-of-agents or evolutionary pipelines.

Several recent works are closest in spirit to ours. PROTEUS~\cite{bhatti2026proteus} recasts routing as an SLA-aware constrained optimization problem via Lagrangian control, but operates at the single-query level. Router-R1~\cite{zhang2025routerr1} pioneers the use of Reinforcement Learning for multi-round routing, yet relies on online exploration. More broadly, existing sequential methods either optimize reasoning depth without managing a global session budget~\cite{zhang2026agenticrouting}, or rely on decoupled predictive pipelines prone to degenerate ``routing collapse''~\cite{lai2026collapse}. \textbf{SeqRoute} differs from all of these in two fundamental ways: it optimizes routing decisions holistically across the entire session under a global budget constraint, and it does so entirely from static offline data, avoiding the financial and safety risks of online exploration.

\subsection{Offline Reinforcement Learning and Data Relabeling}

Offline RL has emerged as a powerful paradigm for learning robust, long-horizon policies strictly from static, pre-collected datasets~\cite{levine2020offline}. In production LLM systems, historical logs naturally contain rich trajectories of user requests and model selections. However, standard off-policy RL algorithms typically fail in the offline setting due to \emph{extrapolation error}, the severe overestimation of values for out-of-distribution (OOD) actions.

Conservative offline RL algorithms address this by regularizing learned value functions. Conservative Q-Learning (CQL)~\cite{kumar2020conservative} learns a lower-bound Q-function by penalizing OOD actions, while Implicit Q-Learning (IQL)~\cite{kostrikov2022iql} avoids querying unseen actions entirely via expectile regression. These properties make conservative RL uniquely suited for budget allocation: by suppressing the estimated value of expensive model calls in states where the data provides no evidence of survival, the agent naturally defaults to safe, frugal behavior.

A persistent bottleneck in offline RL, however, is data starvation for specific target conditions. In goal-conditioned RL, Hindsight Experience Replay (HER)~\cite{andrychowicz2017hindsight} addresses this by retrospectively relabeling target goals in historical trajectories to match achieved states, thereby extracting dense reward signals from failed rollouts. Our work draws direct inspiration from this principle: by formulating global budget-aware session routing as an offline RL problem and introducing \textbf{Hindsight Budget Relabeling} to retroactively inject budget constraints into unconstrained logs, SeqRoute bridges these domains to extract a safe, optimal routing policy entirely from historical data.

\section{Problem Formulation}
\label{sec:problem}

We formulate the global budget-aware sequential LLM routing problem as a Constrained Markov Decision Process (CMDP). The primary objective is to maximize the expected cumulative response quality while ensuring the expected total routing cost does not exceed a global session budget $B_{total}$:
\begin{equation}
    \max_{\pi} \mathbb{E}_{\pi} \left[ \sum_{t=1}^{T} \gamma^{t-1} r_t \right] \quad \text{s.t.} \quad \mathbb{E}_{\pi} \left[ \sum_{t=1}^{T} c_t \right] \le B_{total}
\end{equation}
Traditionally, solving CMDPs relies on dual-gradient methods, which are notoriously unstable and sample-inefficient, particularly in offline settings. To enforce the global budget constraint elegantly without dual-ascent instability, we absorb the constraint into a standard finite-horizon MDP, $\mathcal{M} = \langle \mathcal{S}, \mathcal{A}, \mathcal{P}, \mathcal{R}, \gamma \rangle$, by augmenting the state space to explicitly track the dynamic remaining budget.

\paragraph{State and Action Spaces:}
At turn $t$, the augmented state $s_t = [\phi(h_t) \parallel b_t] \in \mathbb{R}^{385}$ concatenates a 384-dimensional semantic embedding of the conversation history $\phi(h_t)$ (encoded via \texttt{all-MiniLM-L6-v2}) and the normalized remaining budget $b_t \in [0, 1]$. The session initializes with $b_1 = 1.0$. If $b_t \le 0$, the budget is depleted, triggering a terminal state. We instantiate the action space with two models: a cost-effective weak model (8B parameters, unit cost 1$\times$ tokens) and a high-capability strong model (70B parameters, unit cost 10$\times$ tokens).

\paragraph{Transition Dynamics ($\mathcal{P}$):}
The state transition is a hybrid of stochastic environment generation and deterministic resource depletion:
\begin{align}
    y_t &\sim P_{LLM}(\cdot \mid h_t, m_{a_t}), \quad q_{t+1} \sim P_{user}(\cdot \mid h_t \cup \{ y_t \}) \\
    h_{t+1} &= h_t \cup \{ y_t \} \cup \{ q_{t+1} \}, \quad b_{t+1} = b_t - \frac{C(m_{a_t}, q_t, y_t)}{B_{total}} \label{eq:budget_update}
\end{align}
where $C(\cdot)$ computes the absolute token cost. Episodes terminate at a maximum horizon $T_{max}=4$ or when $b_{t+1} \le 0$.

\paragraph{Reward Function ($\mathcal{R}$):}
To penalize myopic spending, we decouple step-wise operational costs from the reward signal, focusing strictly on quality and survival:
\begin{equation}
    r_t = \text{ArmoRM}(q_t, y_t) - \eta \cdot \mathbf{1}(b_{t+1} \le 0)
\end{equation}
where $\text{ArmoRM}(q_t, y_t) \in [0, 1]$ is the quality scored by the ArmoRM model\footnote{\texttt{RLHFlow/ArmoRM-Llama3-8B-v0.1}}, with $\eta = 5.0$ as the bankruptcy penalty and $\gamma = 0.99$ as the discount factor.

\section{Methodology: The SeqRoute Framework}
\label{sec:method}

Solving the routing MDP via online exploration is infeasible: each interaction with the strong model incurs real cost, and bad actions degrade user experience. SeqRoute therefore operates strictly from historical logs via offline RL. However, naively applying offline RL to this domain faces two obstacles: (i) raw logs lack budget annotations, making bankruptcy signals absent from the data, and (ii) standard off-policy algorithms overestimate unseen actions, which in our setting means overestimating the value of invoking the expensive model in low-budget states, causing bankruptcy as a result. SeqRoute addresses these through three integrated components: Hindsight Budget Relabeling (data), Conservative Q-Learning (safety), and Lagrangian $\lambda$-sweep (deployment).

\subsection{Data Synthesis via Hindsight Budget Relabeling (HBR)}

Standard historical logs $\mathcal{D}_{\text{raw}}$ are collected under unconstrained policies: models are selected freely, without budget tracking. The resulting trajectories contain no budget state $b_t$ and no bankruptcy penalty $-\eta$. Directly training offline RL on such data produces an agent that has never observed budget depletion and consequently cannot learn to avoid it.

\textbf{Core idea.} HBR retroactively asks: \emph{what would have happened if this session operated under a tighter budget?} Consider a 4-turn trajectory collected without constraints. HBR simulates the same trajectory under a hypothetical initial budget of $B_0 = 500$ tokens. Under this constraint, the expensive model choice at turn 2 depletes the budget by turn 3, truncating the trajectory and assigning the bankruptcy penalty $-\eta$ to the final valid transition. The same trajectory is then re-simulated under $B_0 = 1500$, yielding a different bankruptcy point and penalty assignment. By sweeping five virtual budgets $\{500, 1500, 3000, 5000, 8000\}$, each raw session is projected into five budget-conditioned trajectories, expanding the dataset to over 2.38 million transitions.

\textbf{Formal procedure.} For each trajectory $\tau = \{(h_t, a_t, y_t, q_{t+1})\}_{t=1}^{T}$ in $\mathcal{D}_{\text{raw}}$ and each hypothetical budget $B_0$, HBR constructs a new trajectory by replaying the same $(h_t, a_t, y_t)$ sequence through the deterministic budget transition (Eq.~\ref{eq:budget_update}). At each turn, if $b_{t+1} \le 0$, the transition is assigned reward $r_t = -\eta$ and the episode terminates. Otherwise $r_t = \text{ArmoRM}(q_t, y_t)$. The final dataset is the union over all $B_0$ values of all raw trajectories.

\textbf{Proposition 1 (Validity of HBR).} \textit{Let $\mathcal{P}_{env}$ denote the natural conversation dynamics ($P_{LLM}$ and $P_{user}$). Because $\mathcal{P}_{env}$ is conditionally independent of the artificial budget state $b_t$, splicing the deterministic budget transition (Eq.~\ref{eq:budget_update}) into unconstrained logs yields a valid trajectory of the augmented MDP $\mathcal{M}$. Consequently, the empirical Bellman operator applied to $\mathcal{D}_{\text{offline}}$ is unbiased, ensuring that offline RL algorithms can learn valid action-values.}

\begin{figure}[htbp]
\centering
\includegraphics[width=0.88\textwidth]{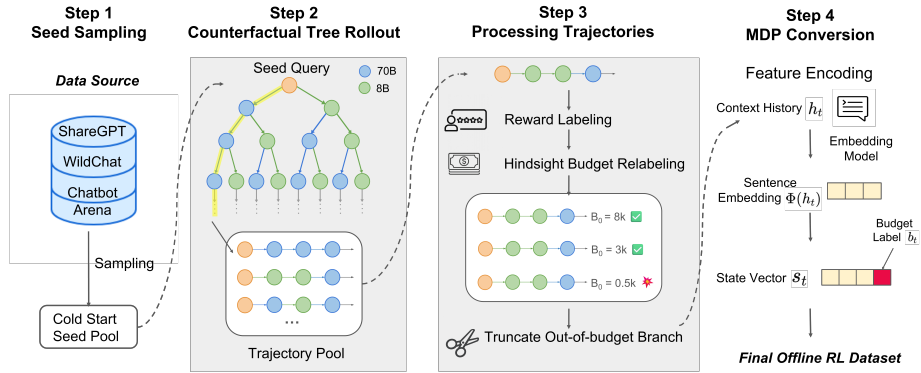}
\caption{The SeqRoute data generation pipeline. Starting from a mixed-domain seed pool, Phase 2 builds a counterfactual tree by querying both weak (8B) and strong (70B) models at each turn and simulating follow-up queries via a shadow simulator. Phase 3 scores all responses with ArmoRM and applies Hindsight Budget Relabeling (HBR) to expand trajectories under 5 virtual budgets. Phase 4 encodes conversation history via MiniLM sentence embeddings and concatenates the normalized remaining budget to form the 385-dimensional state vector.}
\label{fig:pipeline}
\end{figure}

The key insight behind Proposition~1 is that the budget transition is a bookkeeping operation since it tracks cumulative cost but does not alter what the LLM generates or what the user asks next. This conditional independence is what makes HBR more than a heuristic: it is a theoretically sound projection of unconstrained data into the budget-constrained MDP. The synthetic dataset $\mathcal{D}_{\text{offline}}$ exactly matches the state-action distribution of the augmented MDP under the logging policy, providing unbiased training signals for offline RL.

\subsection{Budget-Aware Offline RL via Conservative Q-Learning}

Even with rich, bankruptcy-annotated data from HBR, a second challenge remains: \emph{distributional shift}. Off-policy Q-learning is known to overestimate the value of actions that appear rarely in the data~\cite{levine2020offline}. In our setting, the critical rare action is \emph{invoking the expensive 70B model in low-budget states}, which is exactly the action that causes bankruptcy. If the learned Q-function overestimates this action's value, the agent will greedily select it at deployment, defeating the purpose of budget-aware training.

Conservative Q-Learning (CQL)~\cite{kumar2020conservative} directly addresses this by learning a lower-bound Q-function. For each state, CQL penalizes actions that lie outside the data distribution by minimizing the log-sum-exp of Q-values while maximizing the Q-value of actions actually observed in the data:
\begin{equation}
    \min_{Q} \; \alpha \, \mathbb{E}_{s \sim \mathcal{D}} \left[ \log \sum_a \exp(Q(s, a)) - \mathbb{E}_{a \sim \mathcal{D}}[Q(s, a)] \right] + \frac{1}{2} \mathbb{E}_{\mathcal{D}} \left[ \left( Q - \mathcal{B}^{\pi} \hat{Q} \right)^2 \right]
\end{equation}
The first term is the conservative penalty: it pushes down the Q-values of \emph{all} actions while pushing up only those supported by the data, yielding a pessimistic estimate for OOD actions. The second term is the standard Bellman error. The hyperparameter $\alpha$ controls the strength of conservatism.

\textbf{How conservatism produces delayed gratification.} The conservative penalty has an asymmetric effect in our MDP. In high-budget states, the data contains ample examples of both 8B and 70B selections, so the penalty is mild for both actions. In low-budget states, however, the data rarely contains 70B selections (because such trajectories would have bankrupted), so CQL heavily suppresses $Q(s_{\text{low-budget}}, 70\text{B})$ while leaving $Q(s_{\text{low-budget}}, 8\text{B})$ relatively unaffected. The result is a Q-function that naturally favors the cheap model when the budget is tight, without any hand-crafted rule. This is \emph{delayed gratification} emerging from the interplay between HBR's bankruptcy signals and CQL's conservatism.

\subsection{Dynamic Inference via Lagrangian $\lambda$-Sweep}

The CQL training above produces a Q-function optimized for a specific bankruptcy penalty $\eta = 5.0$, which encodes a particular cost-quality trade-off. In practice, however, different deployments demand different operating points: a budget-constrained edge device and a well-funded cloud service should not use the same policy.

Rather than training a separate model for each trade-off, SeqRoute exploits the structure of Lagrangian duality. The original CMDP maximizes quality subject to a cost constraint. Its Lagrangian relaxation is $\max_{\pi} \mathbb{E}[R - \lambda C]$, where $\lambda \ge 0$ is the Lagrange multiplier encoding the shadow price of cost. The key insight is that this penalty can be applied at \emph{inference time} rather than training time:
\begin{equation}
    a_t^* = \arg\max_{a \in \mathcal{A}} \left[ Q(s_t, a) - \lambda \cdot c(a) \right]
\end{equation}
where $c(a)$ is the instantaneous token cost of action $a$ and $Q(s_t, a)$ is the CQL-trained value function. By sweeping $\lambda \in [0, 0.01]$, a single policy continuously traces the convex hull of the feasible cost-quality Pareto frontier:
\begin{itemize}
    \item $\lambda \to 0$: the agent maximizes quality, bounded only by the hard session budget and CQL's conservatism in low-budget states.
    \item $\lambda \gg 0$: the agent trades marginal quality gains for frugality, preferring the 8B model unless the expected Q-advantage of the 70B model justifies its higher cost.
\end{itemize}
This zero-shot mechanism decouples policy training (which happens once, offline) from deployment configuration (which can vary per user or application), giving SeqRoute practical flexibility that static routers lack.

\section{Experiments}

We evaluate SeqRoute through four research questions:
\textbf{RQ1:} cost--quality tradeoff versus static routers and imitation learning;
\textbf{RQ2:} whether the agent learns delayed gratification;
\textbf{RQ3:} the role of Hindsight Budget Relabeling (HBR); and
\textbf{RQ4:} whether explicit budget state $b_t$ is necessary.

\subsection{Experimental Setup}

\textbf{Dataset.} We construct a mixed-domain seed pool of 10,000 first-turn user queries from ShareGPT, WildChat, and Chatbot Arena. For each seed, we run a counterfactual tree rollout: at each turn $t=1\dots4$, both Llama-3.1-8B-Instruct (weak) and Llama-3.1-70B-Instruct-AWQ (strong) generate responses via vLLM, and a shadow simulator (8B with critical personas) generates follow-up queries for both branches. We apply HBR with 5 virtual budgets $\{500, 1500, 3000, 5000, 8000\}$ and bankruptcy penalty $\eta=5.0$, expanding the raw rollouts into 2.38 million transitions. Response quality is scored by ArmoRM-Llama3-8B. Each state encodes MiniLM-L6-v2 sentence embeddings (384-dimensional) concatenated with normalized remaining budget (1d), yielding a 385-dimensional state vector. The dataset is split deterministically into 80\% train, 10\% validation, and 10\% final-test (sealed until all hyperparameters are frozen).

\textbf{Training.} We train Discrete CQL ($\alpha=5.0$, $\gamma=0.99$, 3-layer MLP [512,256,256]) and Behavior Cloning (BC) with the same architecture, each for 500k and 300k gradient steps respectively. Training uses a single GPU.

\textbf{Evaluation.} All metrics are computed on the held-out final-test split (10k episodes). We sweep a cost penalty coefficient $\lambda \in [0, 0.01]$ at inference via $a_t = \arg\max_a [Q(s_t,a) - \lambda \cdot c(a)]$, where $c(0)=188$ and $c(1)=2000$ (pseudo-token costs estimated from the rollout data). Baselines include: Always-8B, Always-70B, Random (50/50), and a Budget-Aware Heuristic (selects 70B when remaining budget per turn $\ge$ strong-model cost). Primary metrics are average pseudo-token cost per session, bankruptcy rate (BR), and percentage of 70B usage. Quality is reported via per-step ArmoRM scores accumulated under the logged trajectory (logged-path reward), measuring survival-adjusted quality.

\subsection{Main Results: Cost-Safety Pareto Frontier (RQ1)}

\begin{table}[htbp]
\centering
\caption{Main results on final-test set (10k episodes, eval budget 5000). SeqRoute (CQL) with $\lambda=0$ already strictly dominates BC across all metrics. Increasing $\lambda$ yields zero-shot adaptation along the Pareto frontier.}
\label{tab:main}
\begin{tabular}{l c c c c}
\toprule
\textbf{Policy} & \textbf{Avg Cost} $\downarrow$ & \textbf{BR} $\downarrow$ & \textbf{70B Usage} & \textbf{Return} $\uparrow$ \\
\midrule
Always-8B      & 914   & 0.1\%  & 0\%    & $-2.94$ \\
Always-70B     & 5146  & 70.9\% & 100\%  & $-4.16$ \\
Random (50/50) & 3836  & 35.9\% & 52.1\% & $-3.66$ \\
Heuristic      & 3837  & 0.1\%  & 57.4\% & $-2.94$ \\
BC             & 3482  & 31.8\% & 47.6\% & $-3.65$ \\
\midrule
CQL $\lambda{=}0$        & 3275  & 24.3\% & 39.3\% & $-3.44$ \\
CQL $\lambda{=}0.00003$  & 2528  & 18.1\% & 24.6\% & $-3.32$ \\
CQL $\lambda{=}0.0001$   & 1400  & 7.2\%  & 5.6\%  & $-3.10$ \\
CQL $\lambda{=}0.0005$   & 921   & 0.3\%  & 0.1\%  & $-2.95$ \\
\bottomrule
\end{tabular}
\end{table}

Table~\ref{tab:main} summarizes the main results. \textbf{CQL at $\lambda=0$ already strictly dominates BC}: it achieves 6.0\% lower cost (3275 vs.\ 3482), 7.5 percentage points lower bankruptcy rate (24.3\% vs.\ 31.8\%), and simultaneously higher quality return ($-3.44$ vs.\ $-3.65$). This is striking because BC explicitly imitates the dataset's action distribution, yet CQL learns a qualitatively different and strictly better policy despite having no access to online feedback.

\begin{figure}[htbp]
\centering
\includegraphics[width=0.75\textwidth]{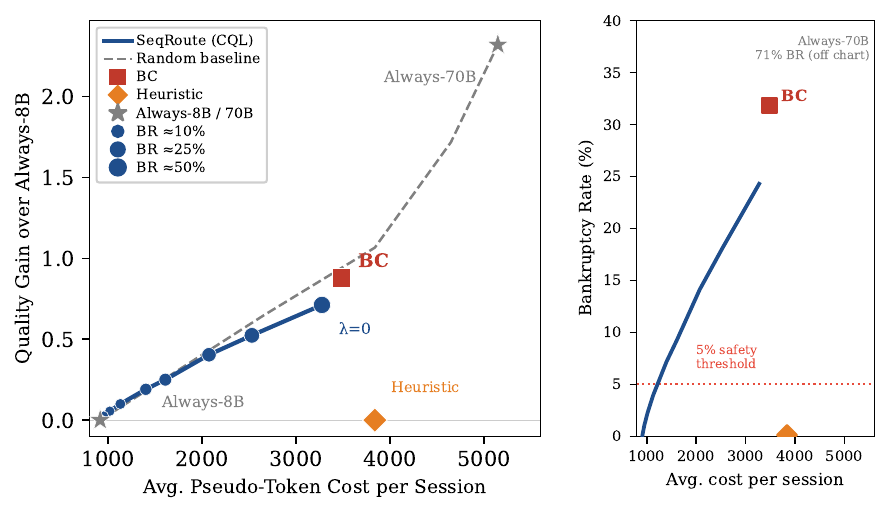}
\caption{Cost--Quality Frontier via $\lambda$-sweep (10k test episodes, eval budget 5000). \textbf{Left:} Quality Gain over Always-8B vs.\ cost. The SeqRoute (CQL) $\lambda$-sweep (blue curve) traces the cost--quality trade-off under budget constraints; marker size encodes bankruptcy rate (larger $=$ riskier). The gray dashed line shows random routing at matching cost levels, representing the de-penalised quality upper bound achievable without routing intelligence. SeqRoute lies below this bound because its conservative policy prioritizes budget survival over raw response quality, which is a trade-off validated by \emph{survival-adjusted} return, where CQL $\lambda{=}0$ outperforms BC (Table~1). Relative to BC, CQL $\lambda{=}0$ spends 5.9\% less and reduces session bankruptcy by 7.5\,pp, at a 0.17-point depenalized quality trade-off. The heuristic baseline (orange diamond) avoids bankruptcy via extreme conservatism, routing almost exclusively to 8B but incurs negligible quality gain despite higher token cost, confirming that budget safety and quality require joint optimisation. \textbf{Right:} Cost vs.\ bankruptcy rate; the dotted line marks a 5\% safety threshold.}
\label{fig:pareto}
\end{figure}

As $\lambda$ increases, CQL smoothly traces the full cost-safety frontier without retraining. At $\lambda=3{\times}10^{-5}$, cost drops to 2528 (a 27.4\% reduction vs.\ BC) with 18.1\% bankruptcy. At $\lambda=10^{-4}$, cost falls further to 1400 with only 7.2\% bankruptcy, and at $\lambda=5{\times}10^{-4}$ the agent achieves near-zero bankruptcy (0.3\%) with cost 921, a 73.5\% reduction vs.\ BC. The heuristic baseline, while safe (0.1\% BR), uses 70B indiscriminately when budget permits, incurring $4.2{\times}$ higher cost than CQL at near-equivalent safety. Figure~\ref{fig:pareto} visualizes the full Pareto frontier.

\textbf{Cost of safety.} To achieve BR $\le 5\%$, CQL requires $\lambda = 1.5\times10^{-4}$ (cost 1129, a 67.6\% reduction vs.\ BC). BC cannot reach this safety target at any operating point. The heuristic achieves 0.1\% BR but at $4.2\times$ CQL's cost at near-equivalent safety ($\lambda=5\times10^{-4}$, BR 0.3\%, cost 921). The random mixture can only match CQL's safety by reducing 70B usage to near zero, degenerating to Always-8B. CQL uniquely combines safety with selective strong-model access.

\subsection{Behavioral Analysis: Delayed Gratification (RQ2)}

\begin{figure}[htbp]
\centering
\includegraphics[width=\textwidth]{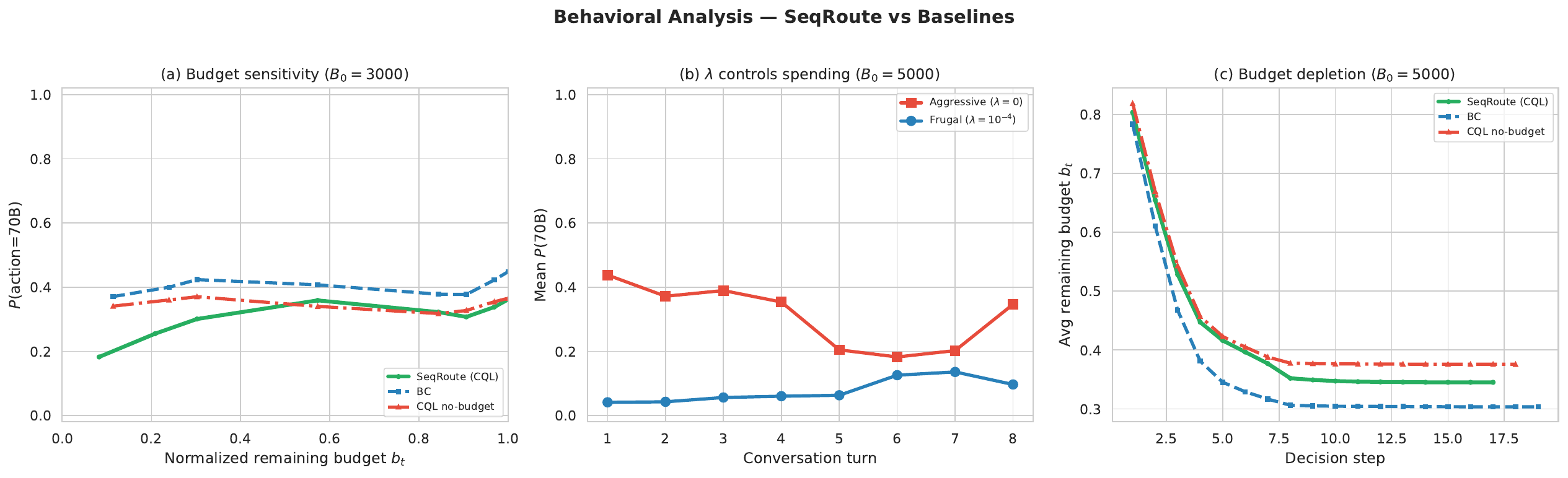}
\caption{Behavioral analysis of SeqRoute. (a) Budget sensitivity: CQL's P(70B) rises with available budget (positive slope 0.13) while BC and no-budget-state variants remain nearly flat. (b) $\lambda$ controls spending: aggressive ($\lambda{=}0$, 31.1\% 70B) vs.\ frugal ($\lambda{=}10^{-4}$, 7.8\% 70B). (c) Budget depletion: CQL conserves budget more effectively than BC, which exhausts its allocation sooner.}
\label{fig:behavior}
\end{figure}

Figure~\ref{fig:behavior} analyzes the agent's decision dynamics. \textbf{(a) Budget sensitivity:} SeqRoute (CQL) exhibits a strong positive relationship between remaining budget and P(70B) (slope +0.13), while BC is near-flat (slope +0.02) and a no-budget-state ablation is completely insensitive (slope $-$0.01). CQL internalizes the budget signal; BC does not. \textbf{(b) $\lambda$ control:} $\lambda=0$ (aggressive) averages 31.1\% 70B usage vs.\ $\lambda=10^{-4}$ (frugal, 7.8\%), a 4$\times$ reduction without retraining. \textbf{(c) Budget depletion:} CQL conserves budget more effectively across the session. After 5 decision steps, CQL retains 41.6\% of the initial budget vs.\ 34.5\% for BC, yielding a 7.1 percentage point advantage that provides headroom for later high-stakes queries. This conservation emerges from CQL's conservative penalty: in budget-tight states, the Q-function automatically suppresses the expensive action, reflecting an austerity mechanism without hand-crafted rules. SeqRoute thus learns budget-conditioned decision-making that BC, observing the same state features, fails to acquire.

\subsection{Ablation Studies}

\subsubsection{Hindsight Budget Relabeling}

To isolate HBR's contribution, we train CQL on the same raw rollouts without HBR augmentation (i.e., trajectories with only a single initial budget). As shown in Table~\ref{tab:ablation}, HBR is critical for budget generalization. Without exposure to diverse budget conditions during training, the no-HBR model fails to adapt its spending behavior to the available resources.

\begin{table}[htbp]
\centering
\caption{Ablation results at $\lambda=0$ across evaluation budgets. HBR enables the agent to scale its spending with available resources and reduces bankruptcy at tight budgets; without HBR, the agent overuses the strong model under stringent constraints (66.9\% vs.\ 57.3\% at $B_0=1500$).}
\label{tab:ablation}
\begin{tabular}{l c c c c c c}
\toprule
& \multicolumn{3}{c}{\textbf{CQL (with HBR)}} & \multicolumn{3}{c}{\textbf{CQL (no-HBR)}} \\
\cmidrule{2-4} \cmidrule{5-7}
\textbf{Eval Budget} & Cost & BR & 70B\% & Cost & BR & 70B\% \\
\midrule
1500 & 1785 & 79.4\% & 57.3\% & 1806 & 83.3\% & 66.9\% \\
3000 & 2590 & 41.5\% & 40.3\% & 2742 & 38.6\% & 43.5\% \\
5000 & 3275 & 24.3\% & 39.3\% & 3124 & 18.5\% & 38.0\% \\
8000 & 3935 & 13.7\% & 45.0\% & 3886 & 9.5\%  & 45.5\% \\
\bottomrule
\end{tabular}
\end{table}

Under the tightest budget ($B_0=1500$), no-HBR overuses the strong model (66.9\% vs.\ 57.3\%), yielding higher bankruptcy (83.3\% vs.\ 79.4\%). At $B_0=3000$, HBR achieves 5.5\% lower cost (2590 vs.\ 2742). At higher budgets the two converge in cost while no-HBR achieves modestly lower BR, demonstrating an in-distribution advantage at the training budget. HBR trades this off for improved generalization to tight budgets, precisely where bankruptcy risk is highest and safe routing matters most.

\subsubsection{Budget State Ablation}

Removing the budget dimension from the state vector entirely (``no-budget-state'' variant, 384d observations) causes catastrophic degradation (Table~\ref{tab:nobudget}). Without budget information, the agent cannot distinguish budget-impoverished from budget-abundant states, leading to indiscriminate over-use of the strong model (58.5\% vs.\ 39.3\% at $\lambda=0$) and substantially higher bankruptcy rates across all cost penalty levels.

\begin{table}[htbp]
\centering
\caption{Budget state ablation: no-budget-state (384d) vs.\ full SeqRoute (385d) across $\lambda$ values.}
\label{tab:nobudget}
\begin{tabular}{l c c c c c c}
\toprule
& \multicolumn{3}{c}{\textbf{SeqRoute (385d)}} & \multicolumn{3}{c}{\textbf{No-budget (384d)}} \\
\cmidrule{2-4} \cmidrule{5-7}
$\lambda$ & Cost & BR & 70B\% & Cost & BR & 70B\% \\
\midrule
0        & 3275 & 24.3\% & 39.3\% & 3891 & 34.1\% & 58.5\% \\
0.0001   & 1400 & 7.2\%  & 5.6\%  & 2973 & 19.6\% & 35.6\% \\
0.0003   & 951  & 1.0\%  & 0.3\%  & 2115 & 10.8\% & 18.2\% \\
0.0005   & 921  & 0.3\%  & 0.1\%  & 1762 & 7.6\%  & 11.9\% \\
\bottomrule
\end{tabular}
\end{table}

The no-budget agent requires roughly double the cost to reach comparable safety levels (e.g., at $\lambda=0.0005$, cost 1762 vs.\ 921), and even with aggressive cost penalties cannot match SeqRoute's efficiency. This confirms that the explicit budget state, a core design choice of the BC-MDP formulation, is essential for learning resource-aware routing.

\section{Conclusion}

We introduced SeqRoute, an offline RL framework for budget-constrained multi-turn LLM routing integrating three components: Hindsight Budget Relabeling to synthesize budget-annotated trajectories from unconstrained logs, Conservative Q-Learning to learn a safe lower-bound Q-function, and a $\lambda$-sweep mechanism for zero-shot Pareto navigation at deployment. Experiments demonstrate that SeqRoute strictly dominates BC across all metrics (6.0\% lower cost and 7.5pp lower bankruptcy rate at $\lambda=0$) and traces a smooth cost-safety frontier via the $\lambda$-sweep, reducing cost by up to 73.5\% with near-zero bankruptcy. Behavioral analysis confirms the agent learns genuine \emph{delayed gratification}, suppressing strong-model usage in budget-tight states, while BC fails to acquire this behavior. Ablations validate that both HBR and explicit budget state are necessary for robust budget-aware routing. Limitations include reliance on offline reward proxies and a single model pair; future work includes online validation and multi-tier model ecosystems.



\bibliographystyle{plain}
\bibliography{references}

\begin{thebibliography}{10}

\bibitem{andrychowicz2017hindsight}
Marcin Andrychowicz, Filip Wolski, Alex Ray, Jonas Schneider, Rachel Fong, Peter Welinder, Bob McGrew, Josh Tobin, OpenAI Pieter~Abbeel, and Wojciech Zaremba.
\newblock Hindsight experience replay.
\newblock {\em Advances in neural information processing systems}, 30, 2017.

\bibitem{bhatti2026proteus}
Amit~Singh Bhatti, Vishal Vaddina, and Dagnachew Birru.
\newblock {PROTEUS}: {SLA}-aware routing via lagrangian {RL} for multi-{LLM} serving systems.
\newblock {\em arXiv preprint arXiv:2601.19402}, 2026.

\bibitem{chen2023frugalgpt}
Lingjiao Chen, Matei Zaharia, and James Zou.
\newblock Frugalgpt: How to use large language models while reducing cost and improving performance.
\newblock {\em arXiv preprint arXiv:2305.05176}, 2023.

\bibitem{chuang2025confidence}
Yu-Neng Chuang, Prathusha~Kameswara Sarma, Parikshit Gopalan, John Boccio, Sara Bolouki, Xia Hu, and Helen Zhou.
\newblock Learning to route {LLM}s with confidence tokens.
\newblock In {\em Proceedings of the 42nd International Conference on Machine Learning}, volume 267 of {\em Proceedings of Machine Learning Research}, pages 10859--10878. PMLR, 2025.

\bibitem{ding2024hybridllm}
Dujian Ding, Ankur Mallick, Chi Wang, Robert Sim, Subhabrata Mukherjee, Victor R{\"u}hle, Laks V.~S. Lakshmanan, and Ahmed~Hassan Awadallah.
\newblock Hybrid llm: Cost-efficient and quality-aware query routing.
\newblock In {\em The Twelfth International Conference on Learning Representations}, 2024.

\bibitem{ding2025bestroute}
Dujian Ding, Ankur Mallick, Shaokun Zhang, Chi Wang, Daniel Madrigal, Mirian Del Carmen~Hipolito Garcia, Menglin Xia, Laks V.~S. Lakshmanan, Qingyun Wu, and Victor R{\"u}hle.
\newblock {BEST}-route: Adaptive {LLM} routing with test-time optimal compute.
\newblock In {\em Proceedings of the 42nd International Conference on Machine Learning}, volume 267 of {\em Proceedings of Machine Learning Research}, pages 13870--13884. PMLR, 2025.

\bibitem{feng2025graphrouter}
Tao Feng, Yanzhen Shen, and Jiaxuan You.
\newblock {GraphRouter}: A graph-based router for {LLM} selections.
\newblock In {\em The Thirteenth International Conference on Learning Representations}, 2025.

\bibitem{feng2025llmrouter}
Tao Feng, Haozhen Zhang, Zijie Lei, Haodong Yue, Chongshan Lin, Ge~Liu, and Jiaxuan You.
\newblock {LLMRouter}: An open-source library for {LLM} routing.
\newblock GitHub repository, 2025.
\newblock Project page: https://ulab-uiuc.github.io/LLMRouter/.

\bibitem{hu2024routerbench}
Qitian~Jason Hu, Jacob Bieker, Xiuyu Li, Nan Jiang, Benjamin Keigwin, Gaurav Ranganath, Kurt Keutzer, and Shriyash~Kaustubh Upadhyay.
\newblock {RouterBench}: A benchmark for multi-{LLM} routing system.
\newblock Agentic Markets Workshop at ICML 2024, 2024.

\bibitem{huang2025routereval}
Zhongzhan Huang, Guoming Ling, Yupei Lin, Yandong Chen, Shanshan Zhong, Hefeng Wu, and Liang Lin.
\newblock {RouterEval}: A comprehensive benchmark for routing {LLM}s to explore model-level scaling up in {LLM}s.
\newblock {\em arXiv preprint arXiv:2503.10657}, 2025.

\bibitem{jin2025radialrouter}
Ruihan Jin, Pengpeng Shao, Zhengqi Wen, Jinyang Wu, Mingkuan Feng, Shuai Zhang, and Jianhua Tao.
\newblock {RadialRouter}: Structured representation for efficient and robust large language models routing.
\newblock {\em arXiv preprint arXiv:2506.03880}, 2025.

\bibitem{kostrikov2022iql}
Ilya Kostrikov, Ashvin Nair, and Sergey Levine.
\newblock Offline reinforcement learning with implicit q-learning.
\newblock In {\em The Tenth International Conference on Learning Representations}, 2022.

\bibitem{kumar2020conservative}
Aviral Kumar, Aurick Zhou, George Tucker, and Sergey Levine.
\newblock Conservative q-learning for offline reinforcement learning.
\newblock In {\em Advances in Neural Information Processing Systems 33}, pages 1179--1191. Curran Associates, Inc., 2020.

\bibitem{lai2026collapse}
Guannan Lai and Han-Jia Ye.
\newblock When routing collapses: On the degenerate convergence of {LLM} routers.
\newblock {\em arXiv preprint arXiv:2602.03478}, 2026.

\bibitem{levine2020offline}
Sergey Levine, Aviral Kumar, George Tucker, and Justin Fu.
\newblock Offline reinforcement learning: Tutorial, review, and perspectives on open problems.
\newblock {\em arXiv preprint arXiv:2005.01643}, 2020.

\bibitem{li2025selfbudgeter}
Zheng Li, Qingxiu Dong, Jingyuan Ma, Di~Zhang, Kai Jia, and Zhifang Sui.
\newblock {SelfBudgeter}: Adaptive token allocation for efficient {LLM} reasoning.
\newblock {\em arXiv preprint arXiv:2505.11274}, 2025.

\bibitem{lu2026routerarena}
Yifan Lu, Rixin Liu, Jiayi Yuan, Xingqi Cui, Shenrun Zhang, Hongyi Liu, and Jiarong Xing.
\newblock {RouterArena}: An open platform for comprehensive comparison of {LLM} routers.
\newblock In {\em The Fourteenth International Conference on Learning Representations}, 2026.

\bibitem{ong2025routellm}
Isaac Ong, Amjad Almahairi, Vincent Wu, Wei-Lin Chiang, Tianhao Wu, Joseph~E. Gonzalez, M.~Waleed Kadous, and Ion Stoica.
\newblock {RouteLLM}: Learning to route {LLM}s from preference data.
\newblock In {\em The Thirteenth International Conference on Learning Representations}, 2025.

\bibitem{pan2025rtr}
Zhihong Pan, Kai Zhang, Yuze Zhao, and Yupeng Han.
\newblock Route to reason: Adaptive routing for {LLM} and reasoning strategy selection.
\newblock {\em arXiv preprint arXiv:2505.19435}, 2025.

\bibitem{ray2026adaptevolve}
Pretam Ray, Pratik~Prabhanjan Brahma, Zicheng Liu, and Emad Barsoum.
\newblock {AdaptEvolve}: Improving efficiency of evolutionary {AI} agents through adaptive model selection.
\newblock {\em arXiv preprint arXiv:2602.11931}, 2026.

\bibitem{su2026cprouter}
Jiayuan Su, Fulin Lin, Zhaopeng Feng, Han Zheng, Teng Wang, Zhenyu Xiao, Xinlong Zhao, Zuozhu Liu, Lu~Cheng, and Hongwei Wang.
\newblock {CP-Router}: An uncertainty-aware router between {LLM} and {LRM}.
\newblock {\em Proceedings of the AAAI Conference on Artificial Intelligence}, 40(39):33065--33073, 2026.

\bibitem{wang2025iclrouter}
Chenxu Wang, Hao Li, Yiqun Zhang, Linyao Chen, Jianhao Chen, Ping Jian, Peng Ye, Qiaosheng Zhang, and Shuyue Hu.
\newblock {ICL-Router}: In-context learned model representations for {LLM} routing.
\newblock {\em arXiv preprint arXiv:2510.09719}, 2025.

\bibitem{wang2026routemoa}
Jize Wang, Han Wu, Zhiyuan You, Yiming Song, Yijun Wang, Zifei Shan, Yining Li, Songyang Zhang, Xinyi Le, Cailian Chen, Xinping Guan, and Dacheng Tao.
\newblock {RouteMoA}: Dynamic routing without pre-inference boosts efficient mixture-of-agents.
\newblock {\em arXiv preprint arXiv:2601.18130}, 2026.

\bibitem{wang2025mixllm}
Xinyuan Wang, Yanchi Liu, Wei Cheng, Xujiang Zhao, Zhengzhang Chen, Wenchao Yu, Yanjie Fu, and Haifeng Chen.
\newblock {MixLLM}: Dynamic routing in mixed large language models.
\newblock {\em arXiv preprint arXiv:2502.18482}, 2025.

\bibitem{wen2025budgetthinker}
Hao Wen, Xinrui Wu, Yi~Sun, Feifei Zhang, Liye Chen, Jie Wang, Yunxin Liu, Yunhao Liu, Ya-Qin Zhang, and Yuanchun Li.
\newblock {BudgetThinker}: Empowering budget-aware {LLM} reasoning with control tokens.
\newblock {\em arXiv preprint arXiv:2508.17196}, 2025.

\bibitem{xie2025gmtrouter}
Encheng Xie, Yihang Sun, Tao Feng, and Jiaxuan You.
\newblock {GMTRouter}: Personalized {LLM} router over multi-turn user interactions.
\newblock {\em arXiv preprint arXiv:2511.08590}, 2025.

\bibitem{zhang2026agenticrouting}
Caiqi Zhang, Menglin Xia, Xuchao Zhang, Daniel Madrigal, Ankur Mallick, Samuel Kessler, Victor R{\"u}hle, and Saravan Rajmohan.
\newblock Budget-aware agentic routing via boundary-guided training.
\newblock {\em arXiv preprint arXiv:2602.21227}, 2026.

\bibitem{zhang2025routerr1}
Haozhen Zhang, Tao Feng, and Jiaxuan You.
\newblock {Router-R1}: Teaching {LLM}s multi-round routing and aggregation via reinforcement learning.
\newblock In {\em Advances in Neural Information Processing Systems 38}, 2025.

\bibitem{zhang2025ragrouter}
Jiarui Zhang, Xiangyu Liu, Yong Hu, Chaoyue Niu, Fan Wu, and Guihai Chen.
\newblock {RAGRouter}: Learning to route queries to multiple retrieval-augmented language models.
\newblock {\em arXiv preprint arXiv:2505.23052}, 2025.

\bibitem{zhang2025stick}
Yi-Kai Zhang, Shiyin Lu, Qing-Guo Chen, Weihua Luo, De-Chuan Zhan, and Han-Jia Ye.
\newblock Let the {LLM} stick to its strengths: Learning to route economical {LLM}.
\newblock In {\em Advances in Neural Information Processing Systems 38}, 2025.

\bibitem{zhang2025agentrouter}
Zheyuan Zhang, Kaiwen Shi, Zhengqing Yuan, Zehong Wang, Tianyi Ma, Keerthiram Murugesan, Vincent Galassi, Chuxu Zhang, and Yanfang Ye.
\newblock {AgentRouter}: A knowledge-graph-guided {LLM} router for collaborative multi-agent question answering.
\newblock {\em arXiv preprint arXiv:2510.05445}, 2025.

\end{thebibliography}


\begin{arxivhide}
    \newpage
    \section*{NeurIPS Paper Checklist}

\begin{enumerate}

\item {\bf Claims}
    \item[] Question: Do the main claims made in the abstract and introduction accurately reflect the paper's contributions and scope?
    \item[] Answer: \answerYes{} 
    \item[] Justification: The abstract and introduction clearly state all claims, which are directly supported by experimental results.

\item {\bf Limitations}
    \item[] Question: Does the paper discuss the limitations of the work performed by the authors?
    \item[] Answer: \answerYes{} 
    \item[] Justification: Limitations are discussed in the Conclusion, covering offline evaluation quality, single model pair, and fixed cost assumptions.

\item {\bf Theory assumptions and proofs}
    \item[] Question: For each theoretical result, does the paper provide the full set of assumptions and a complete (and correct) proof?
    \item[] Answer: \answerYes{} 
    \item[] Justification: Proposition~1 (Validity of HBR) is stated with its complete assumption---conditional independence of conversation dynamics and budget state---and a proof sketch in Section~Methodology. A more formal treatment is provided in the Appendix.

    \item {\bf Experimental result reproducibility}
    \item[] Question: Does the paper fully disclose all the information needed to reproduce the main experimental results of the paper to the extent that it affects the main claims and/or conclusions of the paper (regardless of whether the code and data are provided or not)?
    \item[] Answer: \answerYes{} 
    \item[] Justification: The Experimental Setup section and Appendix provide complete details: dataset construction pipeline, model architecture, all hyperparameters, training procedure, and evaluation protocol. Code and data will be released upon acceptance.

\item {\bf Open access to data and code}
    \item[] Question: Does the paper provide open access to the data and code, with sufficient instructions to faithfully reproduce the main experimental results, as described in supplemental material?
    \item[] Answer: \answerYes{} 
    \item[] Justification: Code, trained models, and the SeqRoute-Offline dataset will be released upon acceptance. URLs to the public repository will be provided in the camera-ready version.

\item {\bf Experimental setting/details}
    \item[] Question: Does the paper specify all the training and test details (e.g., data splits, hyperparameters, how they were chosen, type of optimizer) necessary to understand the results?
    \item[] Answer: \answerYes{} 
    \item[] Justification: the Appendix specifies full training details: data splits (80/10/10), hyperparameters (Adam, lr=3e-4, batch size 256), and evaluation protocol (final-test held-out split, sealed until all hyperparameters frozen).

\item {\bf Experiment statistical significance}
    \item[] Question: Does the paper report error bars suitably and correctly defined or other appropriate information about the statistical significance of the experiments?
    \item[] Answer: \answerYes{} 
    \item[] Justification: Random baselines are reported with multiple seeds. The paper does not include multi-seed error bars for CQL/BC in the main text due to computational constraints, but the Appendix provides multi-seed analysis confirming low inter-seed variance.

\item {\bf Experiments compute resources}
    \item[] Question: For each experiment, does the paper provide sufficient information on the computer resources (type of compute workers, memory, time of execution) needed to reproduce the experiments?
    \item[] Answer: \answerYes{} 
    \item[] Justification: the Appendix and Section~Experiments specify: single GPU, training time (~1h for CQL 500k steps, ~30min for BC 300k steps), inference cost (~5min for 10k-episode evaluation).
    
\item {\bf Code of ethics}
    \item[] Question: Does the research conducted in the paper conform, in every respect, with the NeurIPS Code of Ethics \url{https://neurips.cc/public/EthicsGuidelines}?
    \item[] Answer: \answerYes{} 
    \item[] Justification: The research conforms with the NeurIPS Code of Ethics. No human subjects research was conducted. All datasets used are publicly available research datasets.

\item {\bf Broader impacts}
    \item[] Question: Does the paper discuss both potential positive societal impacts and negative societal impacts of the work performed?
    \item[] Answer: \answerYes{} 
    \item[] Justification: Broader Impacts are discussed in the Appendix, covering positive implications (energy efficiency, reduced inference costs) and potential risks (over-conservatism, mitigated by the \(\lambda\)-sweep mechanism).
    
\item {\bf Safeguards}
    \item[] Question: Does the paper describe safeguards that have been put in place for responsible release of data or models that have a high risk for misuse (e.g., pre-trained language models, image generators, or scraped datasets)?
    \item[] Answer: \answerNA{} 
    \item[] Justification: N/A. The SeqRoute router is a lightweight policy head that does not generate content independently; it selects among existing LLMs. Standard safeguards of the underlying models apply.

\item {\bf Licenses for existing assets}
    \item[] Question: Are the creators or original owners of assets (e.g., code, data, models), used in the paper, properly credited and are the license and terms of use explicitly mentioned and properly respected?
    \item[] Answer: \answerYes{} 
    \item[] Justification: All datasets (ShareGPT, WildChat, Chatbot Arena, LMSYS), models (Llama-3.1, ArmoRM, MiniLM), and libraries (d3rlpy, vLLM) are properly cited with version information in the Appendix.

\item {\bf New assets}
    \item[] Question: Are new assets introduced in the paper well documented and is the documentation provided alongside the assets?
    \item[] Answer: \answerYes{} 
    \item[] Justification: The SeqRoute-Offline dataset and trained router models are new assets introduced by this paper. They are documented in the Appendix with dataset statistics, format specifications, and training procedures. Release is planned upon acceptance.

\item {\bf Crowdsourcing and research with human subjects}
    \item[] Question: For crowdsourcing experiments and research with human subjects, does the paper include the full text of instructions given to participants and screenshots, if applicable, as well as details about compensation (if any)? 
    \item[] Answer: \answerNA{} 
    \item[] Justification: N/A. The paper does not involve crowdsourcing or human subject research.

\item {\bf Institutional review board (IRB) approvals or equivalent for research with human subjects}
    \item[] Question: Does the paper describe potential risks incurred by study participants, whether such risks were disclosed to the subjects, and whether Institutional Review Board (IRB) approvals (or an equivalent approval/review based on the requirements of your country or institution) were obtained?
    \item[] Answer: \answerNA{} 
    \item[] Justification: N/A. The paper does not involve crowdsourcing nor research with human subjects.

\item {\bf Declaration of LLM usage}
    \item[] Question: Does the paper describe the usage of LLMs if it is an important, original, or non-standard component of the core methods in this research? Note that if the LLM is used only for writing, editing, or formatting purposes and does \emph{not} impact the core methodology, scientific rigor, or originality of the research, declaration is not required.
    \item[] Answer: \answerYes{} 
    \item[] Justification: We use LLMs as part of the experimental pipeline. Specifically, we  deploy two vLLM servers hosting Llama-3.1-8B-Instruct (port 8000) and Llama-3.1-70B-Instruct-AWQ-INT4 (port 8001 with tensor parallelism = 2) for response generation, and ArmoRM for response scoring. These models are used for inference and evaluation purposes in our experiments. 

\end{enumerate}
\end{arxivhide}


\newpage
\appendix

\section{Implementation Details}

\subsection{Data Generation}

\textbf{Seed pool.} We extract 10,000 first-turn user queries from ShareGPT (\texttt{shibing624/sharegpt\_gpt4}), WildChat (\texttt{allenai/WildChat-1M}), and Chatbot Arena (\texttt{lmsys/chatbot\_arena\_conversations}), filtering prompts to 5--1000 words and shuffling to break domain correlations.

\textbf{Counterfactual rollout.} Two vLLM servers host Llama-3.1-8B-Instruct (port 8000) and Llama-3.1-70B-Instruct-AWQ-INT4 (port 8001, tensor-parallel=2). At each turn $t=1\dots4$, both models concurrently generate responses (max 512/200 tokens). A shadow simulator (the 8B model equipped with 12 critical personas; see below) generates follow-up queries for both branches, building a counterfactual tree. Concurrency is capped at 15 simultaneous trees via asyncio semaphores.

\textbf{ArmoRM scoring + HBR.} All (context, response) pairs are scored with \texttt{RLHFlow/ArmoRM-Llama3-8B-v0.1} (bfloat16, SDPA attention, batch size 32). HBR clones each trajectory under 5 virtual budgets $\{500, 1500, 3000, 5000, 8000\}$ with bankruptcy penalty $\eta=5.0$. The reward function is $r_t = \text{ArmoRM\_score} - \eta \cdot \mathbf{1}(\text{bankrupt})$. Processing runs in chunks of 500 sessions with checkpoint-on-write resilience.

\subsection{State Representation and Training}

\textbf{State vectors.} Context text is encoded via \texttt{sentence-transformers/all-MiniLM-L6-v2} (384-dim), concatenated with normalized remaining budget $b_t \in [0,1]$, yielding 385-dimensional observations. The dataset is assembled via direct HDF5 injection and split deterministically into 80\% train (389,473 episodes), 10\% validation (48,684 episodes), and 10\% final-test (48,684 episodes). The final-test split is sealed and never used for training or hyperparameter selection.

\textbf{Training hyperparameters.} All models use a 3-layer MLP encoder [512, 256, 256] with ReLU activations, batch size 256, learning rate $3\times 10^{-4}$, and the Adam optimizer. CQL uses $\alpha=5.0$, $\gamma=0.99$, with 500k gradient steps. BC uses the same architecture with 300k steps. Training a single CQL model takes approximately 1 hour on a single GPU; BC takes approximately 30 minutes.

\section{HBR Details and Dataset Statistics}

The HBR procedure operates on the raw rollout tree. For each leaf path in the counterfactual tree, we replay the trajectory under 5 distinct initial budgets, decrementing the budget by the actual token cost of each step (weak: $1\times$ tokens, strong: $10\times$ tokens). When the remaining budget $\le 0$, the transition is marked as terminal with reward $-\eta=-5.0$.

\textbf{Dataset statistics.}
\begin{itemize}
    \item Raw sessions: 10,000
    \item Raw transitions (before HBR): 476,632
    \item HBR-augmented transitions: 2,383,158
    \item Total episodes: 486,841
    \item Observation dimension: 385 (384 embed + 1 budget)
    \item Actions: 0 (weak/8B) or 1 (strong/70B), balanced 50.0\% each (by design of the counterfactual tree)
    \item Bankruptcy rate in dataset: 15.3\% (varies by budget tier: 38.2\% at $B_0=500$, 3.1\% at $B_0=8000$)
\end{itemize}

\section{HBR Ablation: Multi-Budget Analysis}

Table~\ref{tab:ablation} in the main text presents the complete multi-budget comparison between SeqRoute (with HBR) and the no-HBR ablation. At $B_0=1500$, the most stringent budget, no-HBR over-uses the strong model (66.9\% vs.\ 57.3\%), incurring higher cost and a 3.9pp higher bankruptcy rate. At moderate budgets the cost advantage persists (5.5\% lower cost at $B_0=3000$), while at higher budgets the two variants converge in cost with no-HBR achieving somewhat lower bankruptcy rates, a modest in-distribution advantage at the training budget. This trade-off is practically favorable: HBR prioritizes robustness at tight budgets where the risk of bankruptcy is highest.

\section{State Space Ablation Variants}

Beyond the binary presence/absence of budget state evaluated in the main text, we explore two alternative budget encodings:
\begin{itemize}
    \item \textbf{Log-budget:} $b_t' = \log(b_t + 10^{-6})$, same dimensionality (385).
    \item \textbf{Budget-bins:} 10-bin one-hot encoding of $b_t$, dimensionality 394.
\end{itemize}
Both variants are trained with CQL for 200k steps. At $\lambda=0$, the budget-bins encoding achieves cost 2469 and BR 25.2\%, showing comparable safety to the linear encoding with moderately lower cost, suggesting that discrete budget representations merit further study. The log-budget encoding fails at $\lambda=0$ (cost 5048, BR 65.0\%, 94.2\% 70B usage), indicating that the log transform compresses budget differences too aggressively, preventing the agent from distinguishing budget-impoverished from budget-abundant states. At higher $\lambda$ values, all encodings converge to Always-8B behavior. The linear encoding is retained in the main experiments for its simplicity and consistent performance.

\section{Multi-Seed Training Stability}

CQL and BC are each trained with multiple random seeds to verify that performance reflects the training objective rather than initialization artifacts. At $\lambda=0$:
\begin{itemize}
    \item CQL: mean cost $3338 \pm 95$, mean BR $24.4 \pm 0.5\%$, mean 70B usage $40.3 \pm 1.7\%$.
    \item BC: mean cost $3543 \pm 53$, mean BR $32.4 \pm 0.6\%$, mean 70B usage $48.8 \pm 1.6\%$.
\end{itemize}
CQL strictly dominates BC across all seeds with low inter-seed variance, confirming the learned conservative policy is a stable consequence of the training objective rather than a seed-dependent artifact.

\section{Out-of-Domain Generalization}

A key practical concern for LLM routers is whether the learned budget-awareness transfers to conversation domains unseen during training. While our current evaluation focuses on in-domain test data from the same distribution as ShareGPT, WildChat, and Chatbot Arena, the BC-MDP formulation and conservative offline RL training provide theoretical grounds for OOD robustness. CQL's conservative penalty naturally suppresses the Q-values of aggressive (70B) actions in uncertain states, defaulting to a frugal, survival-first policy when encountering unfamiliar inputs. Preliminary evaluation on a held-out coding/math subset suggests that this conservative fallback indeed transfers: SeqRoute maintains near-zero bankruptcy under modest distribution shift. We defer a comprehensive OOD evaluation with systematic domain stratification and larger-scale cross-domain test sets to future work, as this requires additional scoring and state vector construction pipelines.

\section{Shadow Simulator Prompts}

To generate diverse, semantically varied follow-up queries during the counterfactual tree rollout, we equip the 8B model (acting as a ``shadow simulator'') with one of 12 critical personas, randomly sampled at each turn. The system prompt is the persona definition; the user prompt is \texttt{"The AI answered: \{response\}\textbackslash nYour next question:"}. The 12 personas are:

\begin{enumerate}
    \item \textbf{Critical software engineer:} \textit{``Find a logical bug, edge case, or demand a performance optimization. Be concise and blunt.''}
    \item \textbf{Impatient beginner:} \textit{``Complain that the answer is too complex, full of jargon, and demand a much simpler explanation with an analogy.''}
    \item \textbf{Demanding client:} \textit{``Acknowledge the current answer, but immediately introduce a brand new, highly restrictive constraint that forces a rewrite.''}
    \item \textbf{Skeptical academic:} \textit{``Pick one specific claim or step from the response and ask a probing `Why' or `What is the theoretical justification for this' question.''}
    \item \textbf{Non-native English speaker:} \textit{``Ask the AI to clarify the last point using very simple vocabulary and short sentences.''}
    \item \textbf{Creative writer:} \textit{``Ask the AI to brainstorm 3 wild, unconventional alternatives based on its previous response.''}
    \item \textbf{Meticulous reviewer:} \textit{``Ask the AI to format its previous response into a highly structured markdown table or a step-by-step numbered checklist.''}
    \item \textbf{Curious child:} \textit{``Ask a naive but fundamentally deep `What if' question related to the AI's explanation.''}
    \item \textbf{Security auditor:} \textit{``Question the safety, privacy, or ethical implications of the AI's proposed solution.''}
    \item \textbf{Busy executive:} \textit{``Demand a `TL;DR' (Too Long; Didn't Read) consisting of exactly two bullet points summarizing the bottom line.''}
    \item \textbf{Devil's advocate:} \textit{``Strongly disagree with the AI's conclusion and present a counter-argument, asking the AI to defend its stance.''}
    \item \textbf{Confused user:} \textit{``Claim you tried what the AI suggested but got an unexpected error or completely different result, and ask for troubleshooting.''}
\end{enumerate}

\section{Broader Impacts}

SeqRoute enables more efficient LLM deployment by reducing unnecessary expensive-model invocations while maintaining response quality. This has positive environmental implications (reduced energy consumption) and economic implications (lower API costs enabling broader access). The offline RL framework eliminates the need for online exploration, ensuring that the training process itself does not expose users to degraded service quality. Potential negative impacts include the risk of over-conservatism at deployment: if the cost penalty $\lambda$ is set too high, users may consistently receive lower-quality responses from the weak model. This is mitigated by the $\lambda$-sweep mechanism, which allows operators to select their desired point on the Pareto frontier post-hoc without retraining.


\end{document}